\documentclass{article}

\usepackage[table,xcdraw]{xcolor}
\usepackage{PRIMEarxiv}
\usepackage{makecell}
\usepackage{svg}
\usepackage[utf8]{inputenc} 
\usepackage[T1]{fontenc} 
\usepackage{hyperref} 
\usepackage{url} 
\usepackage{booktabs} 
\usepackage{amsfonts} 
\usepackage{nicefrac} 
\usepackage{microtype} 
\usepackage{lipsum}
\usepackage{fancyhdr} 
\usepackage{graphicx} 
\usepackage{caption}
\usepackage{subcaption}
\graphicspath{{media/}} 
\usepackage{authblk}
\usepackage[natbibapa]{apacite}
\title{Federated learning in food research}

\author[1]{Zuzanna Fendor}
\author[1]{Bas H.M. van der Velden}
\author[1]{Xinxin Wang}
\author[1]{Andrea Jr. Carnoli}
\author[1]{Osman Mutlu}
\author[1]{Ali Hürriyetoğlu}
\affil[1]{Wageningen Food Safety Research (WFSR), part of Wageningen University \& Research, Akkermaalsbos 2, 6708 WB Wageningen, The Netherlands}

\begin{document}
\maketitle

\begin{abstract}
Research in the food domain is at times limited due to data sharing obstacles, such as data ownership, privacy requirements, and regulations. While important, these obstacles can restrict data-driven methods such as machine learning. Federated learning, the approach of training models on locally kept data and only sharing the learned parameters, is a potential technique to alleviate data sharing obstacles. This systematic review investigates the use of federated learning within the food domain, structures included papers in a federated learning framework, highlights knowledge gaps, and discusses potential applications. A total of 41 papers were included in the review. The current applications include solutions to water and milk quality assessment, cybersecurity of water processing, pesticide residue risk analysis, weed detection, and fraud detection, focusing on centralized horizontal federated learning. One of the gaps found was the lack of vertical or transfer federated learning and decentralized architectures.

\end{abstract}


\section{Introduction}
Lack of access to safe and nutritious food is detrimental to the health and wellbeing of people. The sustainable development goal aims to end hunger, achieve food security, improve nutrition and promote sustainable agriculture (Goal 2 | Department of Economic and Social Affairs, n.d.). One approach to contribute to this goal, is to use machine learning (ML) models to gain insight in current issues, help to predict future scenarios, and anticipate on those scenarios. 
\par
ML models are trained directly on data to perform a task, such as classifying the content of an image. For instance, a model can be trained to determine whether the fruit in an image is a pear or an apple. The model learns by making adjustments to its parameters based on the data to improve its performance. A lot of examples are needed to reflect the variety of scenarios and train a reliable ML model. However, when the data originates from just one source, a model might have a limited view of reality. For instance, an apple orchard might have access to many examples of apples, but just a few examples of pears, making the model better at recognizing apples rather than pears. Thus, data sharing between different data sources is beneficial to the size and variety of data sets and to the model performance.
\par
However data sharing, especially in case of private data, becomes a challenge when multiple data owners are involved. Data sharing can be challenging due to legal restrictions and regulations concerning privacy, technical limitations, and data owners being competitors. There is a lot of attention to this challenge in medicine, where data sharing is difficult due to patient privacy requirements \citep{kaissis_secure_2020}. The same need for secure data sharing solutions is also present in the food domain, like in food safety. Here, data owners are reluctant to share their data due to fear for liability, bad publicity, or loss of business advantage over competitors \citep{qian_perspective_2022}.
\par
One way to solve the above-mentioned limitation is via Federated Learning (FL), a type of privacy-preserving ML implemented to train a model without revealing the data themselves \citep{mcmahan_communication-efficient_2023,rieke_future_2020,kaissis_secure_2020}. Within FL the data and the model training are local (i.e. do not leave the data’s owner storage), and only the model parameters are shared to make the model to improve \citep{mcmahan_communication-efficient_2023}. 
\par
Food research often involves scattered data from various research facilities or competing commercial producers reluctant to share their information. Therefore, privacy-preserving ML methods, like FL, can be very useful for this field. Surprisingly, we have not found a review of FL in food research, while extensive reviews of FL applications have been published in other fields, like agriculture, smart cities, air pollution monitoring, and medicine \citep{ilic_federated_2023,jiang_federated_2020,li_review_2020,nguyen_federated_2022,xu_federated_2021,zalik_review_2023}.). This work fills the gap by summarizing current achievements, applications, and reasons for the use of FL in food research, and highlighting encountered pitfalls. 
\par
The main research questions we aim to answer here are: 
\begin{itemize}
 \item “What food research problems are currently being solved by federated learning?”
 \item What are the gaps of federated learning use in food research?
 
\end{itemize}

This paper is structured as followed. In section 2, we will define a federated learning framework. Section 3 provides an overview of the current application space of federated learning in food and categorizes all included papers into this framework. In section 4, we will give an outlook for future use of federated learning in the food domain.

\section{Federated learning framework}
The core principle of FL is uniquely defined – as training the model on locally stored data and only sharing the model parameters – but the related implementation varies with the challenge at hand. We categorize the essential variants of federated learning through the federated learning framework. In this section we will explain federated learning and present the framework of federated learning used to categorize the encountered literature.
\par
FL was introduced as “a decentralized approach of leaving the data distributed on the mobile devices, and learning a shared model by locally-computed updates’’ \citep{mcmahan_communication-efficient_2023}. Following this definition FL is applied when each computer (i.e., client node) possess and does not share their own data. In the FL context, each client node trains an ML model on their local and private data. Then they share the model parameters, which are aggregated to yield a global model. Such an aggregation step can be performed using different strategies, the simplest of which is averaging. Finally, each local model is updated with the parameters of the new global model. The process of sending the global model to the client nodes, training, and aggregating the results repeats until a maximum number of repetitions or a predefined performance minimum (e.g., model accuracy) is reached. 
\par
Nowadays FL is not anymore restricted to the definition of \cite{mcmahan_communication-efficient_2023}, but it has been expanded to other scenarios \citep{huang_cross-silo_2022,kairouz_advances_2021,liu_vertical_2023,yang_federated_2019,yuan_decentralized_2023,zhang_survey_2021}. We will categorize FL approaches by type of server architecture, the clients, and data partitioning to form a framework. Server architecture is divided into centralized and decentralized \citep{yuan_decentralized_2023}. The type of client nodes is divided into cross-device and cross-silo \citep{huang_cross-silo_2022,kairouz_advances_2021}. Data partitioning amongst clients is divided into horizontal learning, vertical learning, and transfer learning. 

\subsection{Server architecture}

\begin{figure}[h!]
\centering
\begin{subfigure}{0.5\textwidth}
 \centering
 \includegraphics[width=.8\linewidth]{ 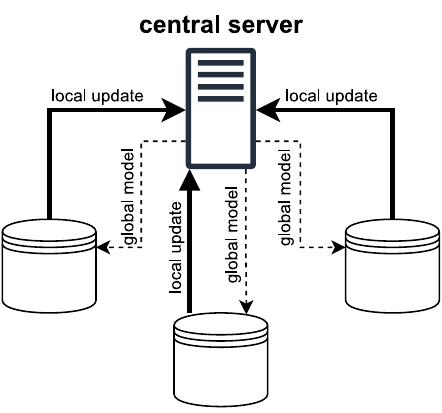}
\end{subfigure}%
\begin{subfigure}{0.5\textwidth}
 \centering
 \includegraphics[width=.8\linewidth]{ 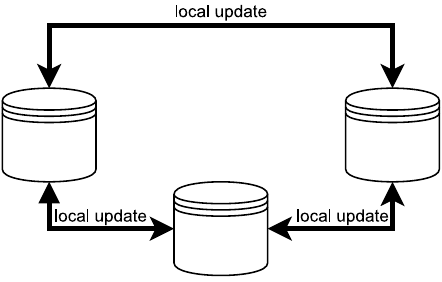}
\end{subfigure}
\caption{Left: a centralized federated learning architecture. Right: a decentralized federated learning architecture}
\label{fig:centr_decent}
\end{figure}

The distinction between centralized and decentralized FL (Figure~\ref{fig:centr_decent}) is determined by the presence of a central server. In centralized FL, a central server is responsible for the aggregation of the results from the client nodes. Such a server is the communication and control center that eases the management and regulation of the FL process. Letting the server manage the training process does not burden the client nodes with both storage and aggregation of the models \citep{yuan_decentralized_2023}. Furthermore, the communication between the central server and the client nodes is generally better regulated and protected from eventual malicious attacks \citep{yuan_decentralized_2023}. Finally, the presence of only one central server makes the FL system easier to operate and manage. Unfortunately, regulating the training process through a single central server introduces a single point of failure i.e., if the central server fails the FL process stops. A server failure leads to a halt in the training process because the local results cannot be aggregated.
\par
Decentralized FL includes many combinations of network topologies and communication protocols such as gossip learning or peer-to-peer learning \citep{hu_decentralized_2019,wink_approach_2021}. In decentralized FL, there is direct communication between the nodes. This means that the nodes share their learned model parameters among each other to continue learning on the received model or to aggregate the parameters from the received model with their own model. Decentralized FL makes the central server obsolete, removing a potential communication bottleneck, a single point of failure and the need of assuming trustfulness of the central server \citep{yuan_decentralized_2023}.

\newpage

\subsection{Type of client nodes}

\begin{figure}[h!]
\centering
\begin{subfigure}{0.5\textwidth}
 \centering
 \includegraphics[width=.8\linewidth]{ 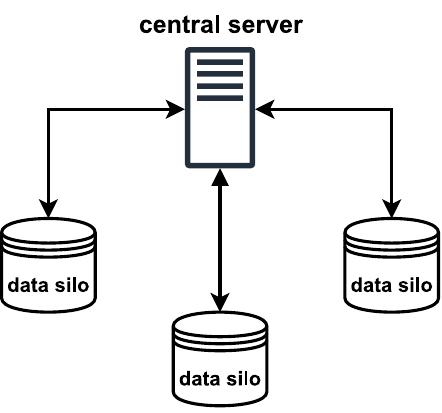}
\end{subfigure}%
\begin{subfigure}{0.5\textwidth}
 \centering
 \includegraphics[width=.8\linewidth]{ 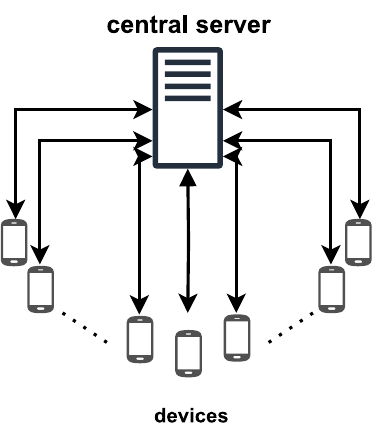}
\end{subfigure}
\caption{Left: Cross-silo federated learning Right: cross-device federated learning.}
\label{fig:nodes}
\end{figure}

The type of client nodes (Figure~\ref{fig:nodes}) can be distinguished in cross-silo and cross-device \citep{kairouz_advances_2021}\citep{kairouz_advances_2021}. This distinction has practical implications to the computational resources and the communication bottlenecks. Cross-silo FL usually refers to different organizations such as research institutions or companies that own a large, isolated data source, i.e., a data silo. The number of client nodes communicating with the central server is limited, which leads to fewer communication bottlenecks. In cross-silo FL, there are usually few carefully selected participants, reducing the risk of a malicious client node trying to disrupt or take unfair advantage of the FL system. With this configuration, there is also less risk of client nodes being unable to participate due to performance requirements, because it is easier to control the quality of the hardware with just few participating client nodes. On the other hand, the presence of slow or unresponsive client nodes has a larger influence on model training, since a participating node being dropped is more drastic in the cross-silo setting. 

\par
Cross-device FL can involve millions of mobile or internet of things (IoT) devices. Consequently, there is less oversight on the participating devices and, in many cases, there will be more variance in the hardware leading to reduced reliability. This type of FL faces larger hardware heterogeneity by design but often offers a large variety and quantity of data.
\par
FL can also be applied to edge computing scenarios. Edge computing is the concept of transferring the data from data sources, such as sensors, IoT devices, or mobile devices, to a geographically proximate edge server \citep{xia_survey_2021}. The edge sever is an intermediate step between performing all training on the data-producing devices and collecting the data form all the devices at one place. FL in edge computing scenarios can be classified as both cross-device and cross-silo depending on the number of participating edge servers and the amount of data stored at each edge server \citep{xia_survey_2021}. In general, we classified all scenarios employing edge servers as cross-device to better reflect the aspect of online data collection from different devices.

\subsection{Data partitioning}

\begin{figure}[h!]
\centering
\includegraphics[width=.6\linewidth]{ 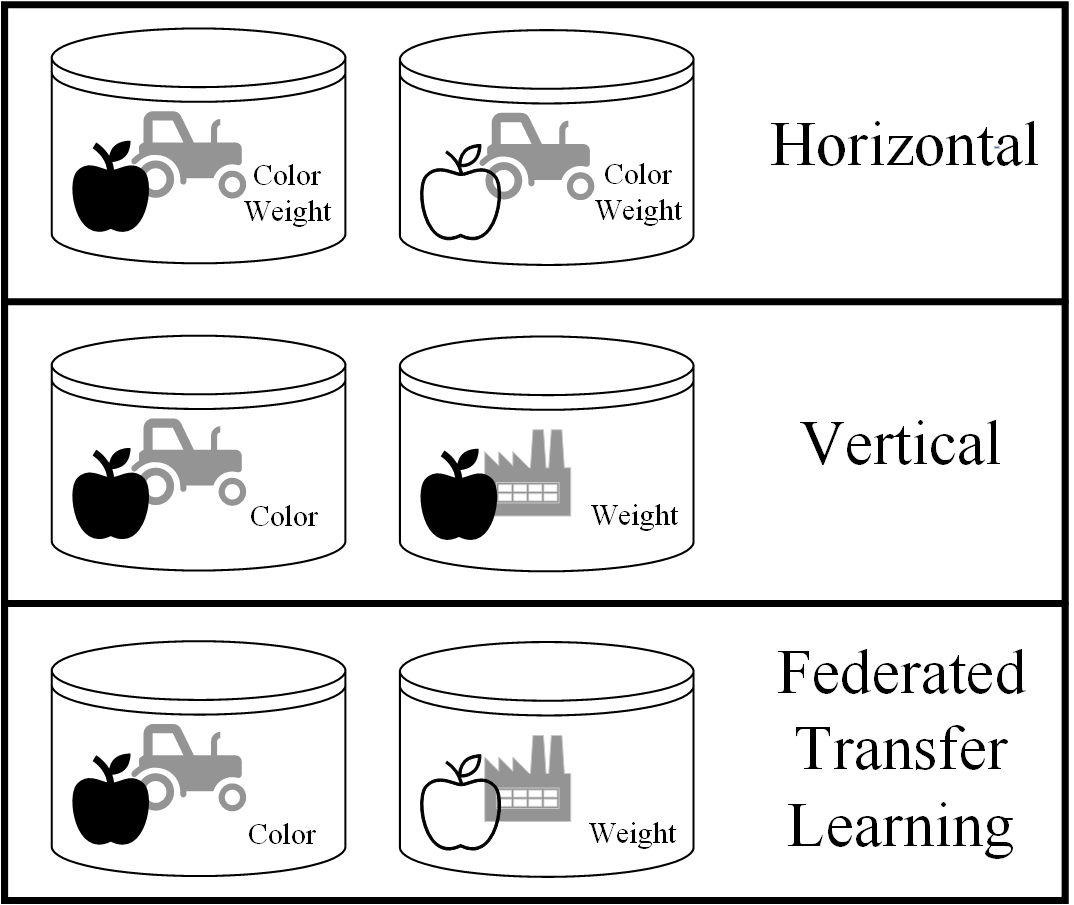}
\caption{Data partitioning in federated learning. Horizontal FL: each local dataset shares features (weight and color) but contains apples from different farms. Vertical FL: The model combines data of the same apple that is matched across datasets. Each dataset contains different features. In this example the features are collected at different sites: a farm and a factory. Federated transfer learning: similar to vertical FL, however there is less overlap between the samples. Instead of entity matching all the samples used for training, the process often involves learning a shared feature space.}
\label{fig:distr}
\end{figure}

Federated learning can also be determined by the way the data is partitioned amongst the nodes, as we also show in Figure ~\ref{fig:distr}. The distinction in data partitioning is important because it determines how the information is spread across the client nodes. We will follow the classification composed of horizontal FL, vertical FL, and federated transfer learning used in the surveys by \cite{yang_federated_2019}, and by \cite{zhang_survey_2021}. In horizontal FL, each node has different samples with the same, or largely similar features. The aim of horizontal FL is to increase the number of training samples with the same set of features. This data partitioning can be applied to both cross-device and cross-silo federated learning \citep{kairouz_advances_2021}. 
\par
In vertical FL there is a large overlap in samples and a smaller overlap in features. This means that the same individual sample is present across different client nodes with different features. Vertical FL is used to increase the feature space of the training data. To illustrate, suppose a shipping company and an apple sauce factory possess information about the same batch of apples. The shipping company owns information about the conditions of the apples during the transport. The factory collects information about the quality of apples, together with descriptive measures such as weight. The two companies can employ vertical FL to collaboratively train a model for the prediction of the quality of apples. Often the label, in this case the quality of apples, is held by one of the client nodes. Thus, the participation of all client nodes is necessary for making model predictions, because the information needed by the model is spread across the client nodes. Furthermore, vertical FL typically assumes a different training protocol from horizontal FL: all parties need to align the individuals in their data, a process called entity alignment \citep{liu_vertical_2023}. Moreover, a single global model is typically not the objective in vertical FL where each client node has their own model that uses their local features. With this FL configuration, the intermediate results from the models of each client are firstly incorporated by the central server into a combined model. Then, the central server assesses the combined model by comparing how well the labels match the predictions. Finally, each client node receives client node-specific feedback from the central server \citep{liu_vertical_2023}. 
\par
The third type of data partitioning is called federated transfer learning. In federated transfer learning, neither the samples nor the features overlap, or they overlap very slightly \citep{liu_secure_2020}. Federated transfer learning aims to improve both the number of samples and the number of features by combining data from different, yet related domains or tasks. Federated transfer learning trains the models in such a way, that the model can be used in the target space using target-specific features after first being trained in the source domain space with other features. This approach leverages the knowledge acquired from one domain to the other \citep{saha_federated_2021}. After the completion of the FL training, the model can be fine-tuned to fit the needs of the client better, while still retaining knowledge gathered from the other related domain. 
\par
Up to our knowledge both vertical FL and federated transfer learning are only possible in the cross-silo setting. These partitioning types require additional communication to accommodate entity matching. Each client node also needs to be identifiable, which is not necessarily the case in the cross-device setting \citep{kairouz_advances_2021}. 
\section{Federated learning in food}
This section shows the literature review findings of FL in food research with extra attention to food safety. Section 3.1 describes the search strategy. Section 3.2 summarizes the found literature in terms of application (section 3.2.1), FL performance (section 3.2.2), reasons for applying FL (3.2.3), and finally the FL framework (section 3.3.4).
\subsection{Search strategy}
An article is eligible if it contains a use case of FL within the food domain. The food domain includes topics such as food safety, yield prediction, improvement and optimization of farming and processing practices, and waste prevention. The minimum requirement for the FL part is that it needs to describe a specific application, as opposed to a brief generic mention like “can be applied to agriculture”. 
\par
The articles were obtained through search on Google Scholar (GS), Scopus, Web of Science (WoS), IEEE Xplore, and CAB abstracts on the 31st of July 2023. The exact search queries are included in the appendix. The search query had two components. The first included the terms: federated learning, federated approach. The second included the terms related to food that can be roughly categorized into supply-chain related terms, adulteration related terms, various individual food items, microbial hazards, chemical hazards, and others. The list of included terms was kept broad to be as complete as possible. 
\par
We limited the search to the title, the abstract and the keywords. The review focuses on “classic” peer-reviewed literature, however encountered “non-classic” literature (i.e., dissertations or preprint articles) that meets the criteria is also included. As concern the publication year, we set it organically to 2016 because it corresponds with the first publication on FL.
\par
The search resulted in a total of 219 unique articles. The articles were checked independently by two researchers for relevance by reading the titles and abstracts. On top of this, the references in relevant articles were scanned for other relevant literature (snowball approach); however this did not lead to the discovery of any earlier unseen publications. 

\begin{figure}[h!]
\centering
 \includegraphics[width=.5\linewidth]{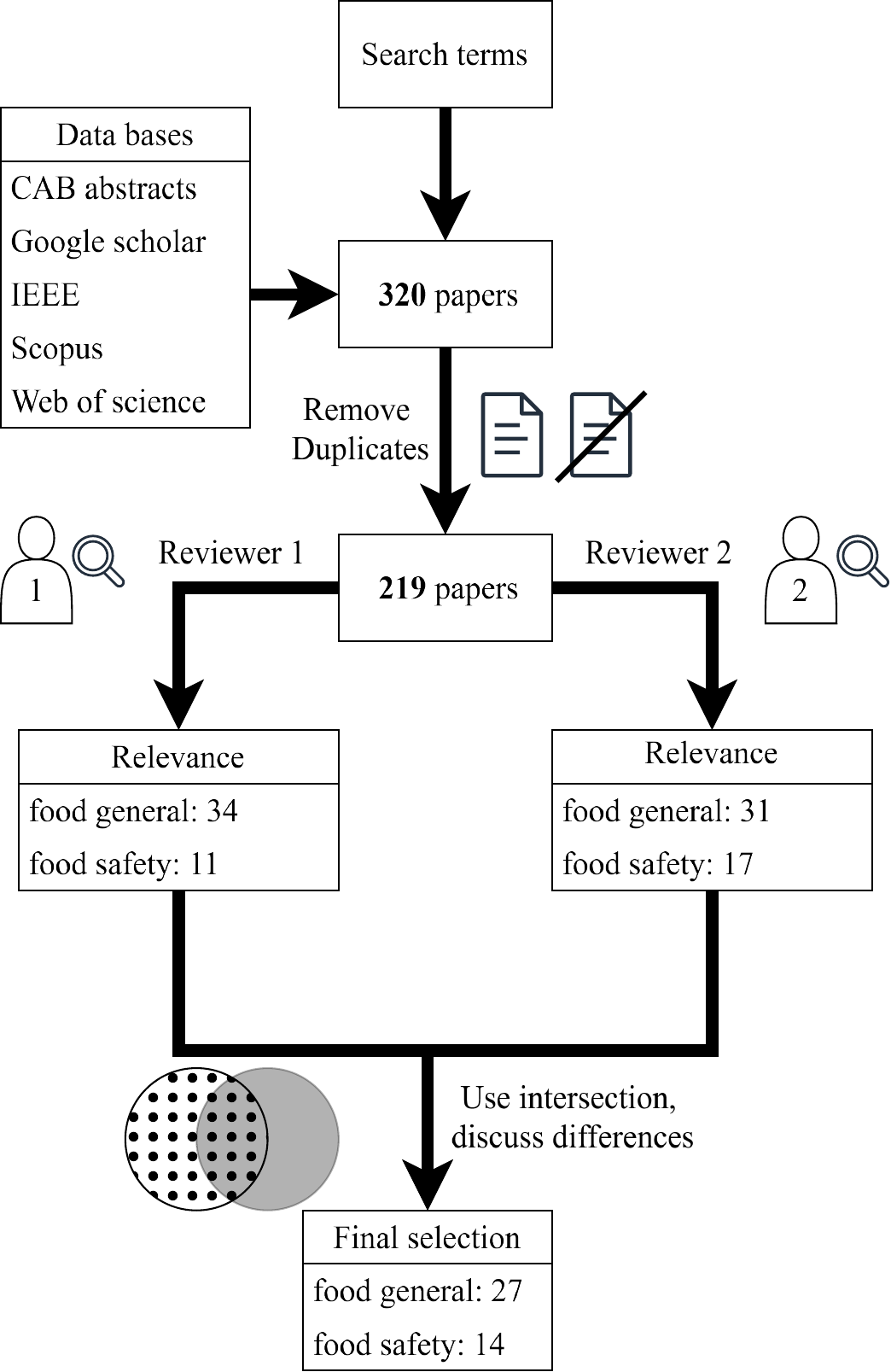}
\caption{Flowchart describing the paper selection and reviewing process. The reviewers looked at the title and the abstract first to assess the relevance. If the relevance could not be determined on the title and abstract alone, the text body was scanned for more information.}
\label{fig:pap_sel}
\end{figure}
\newpage
\subsection{Literature findings}
We arrived at 41 distinct articles that were relevant to the topic of FL in food. Out of those 41 articles, 14 were about food safety specifically. The entire reviewing process can be also viewed in Figure~\ref{fig:pap_sel}. 
\subsubsection{Application overview}
This section provides an overview of the use of federated learning in the food domain, by summarizing the found literature on the topic. The overview is condensed in Table~\ref{table:domainFood}. 
\par
\begin{table}[]
\resizebox{\textwidth}{!}{%
\begin{tabular}{
>{\columncolor[HTML]{FFFFFF}}l 
>{\columncolor[HTML]{FFFFFF}}l 
>{\columncolor[HTML]{FFFFFF}}l 
>{\columncolor[HTML]{FFFFFF}}l }
\textbf{Domain}               & \textbf{Publication}            & \textbf{Data type}       & \textbf{Data content}                           \\
Yield prediction          & \cite{durrant_role_2022}       & Tabular        & Sensor data (weather soil)                    \\
                  &                  & Imaging        & Satellite  images                        \\
                  & \cite{manoj_federated_2022}        & Tabular        & Sensor data                            \\
                  & \cite{zhang_maize_2023}     & Tabular        & Breeding  test phenotypes, environmental meteorology       \\
                  & \cite{bera_e-cropreco_2023}        & Tabular        & Weather, soil, environment sensors                \\
                  & \cite{idoje_federated_2023}       & Tabular        & Climatic  data                          \\
Process monitoring and optimization & \cite{gengler_challenges_2019}          & Tabular        & Milk content, behavior, genome data                \\
                  & \cite{huang_high-precision_2022}     & Imaging        & Chicken  images                         \\
                  & \cite{mao_fedaar_2022}         & Tabular        & Sensor data                            \\
                  & \cite{bowler_domain_2021}      & Tabular  time series & Ultrasonic  sensor features                   \\
                  & \cite{cui_boosted_2022}        & Tabular        & Irrigation demand, evapotranspiration, infiltration from cropland \\
                  & \cite{elhachmi_federated_2022}   & Tabular  time series & Water  consumption                        \\
                  & \cite{pei_fed-iout_2023}        & Tabular        & Sensor data                            \\
Crop disease            & \cite{aggarwal_contemporary_2022}     & Imaging,  tabular  & Unspecified                            \\
                  & \cite{mehta_empowering_2023}  & Imaging        & Wheat images                           \\
                  &\cite{mehta_revolutionizing_2023} & Imaging        & Maize  images                          \\
                  & \cite{yu_energy-aware_2022}        & Imaging        & UAV-gathered images                        \\
                  & \cite{deng_multiple_2022}       & Imaging        & Orchard  and apple images                    \\
                  & \cite{tong_detection_2022}        & Imaging        & Fruit images                           \\
                  & \cite{khan_federated_2022}       & Imaging        & UAV-gathered  images                       \\
Food image classification      & \cite{hsieh_bicameralism_2019}       & Imaging        & Food images                            \\
                  & \cite{polap_hybridization_2022}     & Imaging        & Fruit  images                          \\
Intrusion detection         & \cite{kumar_pefl_2022}      & Tabular time series  & Sensor data                            \\
                  & \cite{jahromi_deep_2021}     & Tabular time series  & Network traffic, sensor, actuator                 \\
                  & \cite{friha_felids_2022}        & Tabular time series  & Traffic data                           \\
                  & \cite{bala_novel_2022}      & Tabular        & Approved  number of blocks, online time, net stake investment  \\
Consumer preference         & \cite{liu_federated_2022}       & Tabular        & Preference scores                        
\end{tabular}%
}
\caption{Domain overview food research applications of federated learning. }
\label{table:domainFood}
\end{table}
\textbf{Yield prediction} 
\par
A good model predicting the yield capacity of crops in certain conditions is not only informative but also helps to take preventive measures. Furthermore, it can help determine the optimal crop variety choice to fit the local conditions. 
\par
Three publications included yield prediction of soybean crops \citep{durrant_role_2022,manoj_federated_2022} and maize crop varieties \citep{zhang_maize_2023}. \cite{durrant_role_2022} used the soybean prediction problem as a use case to demonstrate horizontal FL and its potential for data sharing in agriculture. The task was to predict the average observed yield per unit area within American counties that posed as data silos. They used sequences of remote satellite images taken before harvest, and tabular data of the weather and soil as two different data-driven scenarios. \cite{manoj_federated_2022} focused on the same task of federated soybean yield prediction and also applied centralized, horizontal FL in a cross-silo setting. Their motivation was to contribute to the agricultural production risk management demands by overcoming data sharing obstacles. \cite{zhang_maize_2023} used vertical, cross-silo FL for the selection of the maize variety based on the optimal yield prediction in the context of collaborative plant breeding. The data was collected from 248 trial sites that posed as client nodes. The varieties of maize were considered to be the overlapping samples and phenotypic traits and environmental conditions were features that could differ across client nodes. The authors applied a federated random forest model that required some central knowledge of the features and samples. The central server received encoded and encrypted aligned feature codes to perform decision-tree-type training.
\par
We found two publications focused on the selection of the correct crop from a wider range of plant species based on the weather and soil conditions \citep{bera_e-cropreco_2023,idoje_federated_2023}. \cite{bera_e-cropreco_2023} considered ten plant species to create a crop prediction model that can be used in an app for crop recommendation called CropReco. \cite{bera_e-cropreco_2023} used an edge server framework and FL was used to enable model training on the edge servers that gathered data directly from the sensor devices, which we consider to be a cross-device setting. \cite{idoje_federated_2023}investigated FL Gaussian Naïve Bayes models that can predict the optimal choice between rice, maize or chickpea crops based on climatic features in a horizontally partitioned dataset. 
\par
\textbf{Process monitoring and optimization}
\par
ML can also be used to monitor and optimize other food production processes. The first application was chicken day-age classification by \cite{huang_high-precision_2022}. Knowing the exact age of a chicken can help to choose the optimal breeding cycle, feeding cycle, and slaughter time of respectively breeding chickens, laying hens, and broilers to reduce costs and environmental pollution. The model took chicken images and trained them in an FL manner on horizontally partitioned local data. Each local model was individualized through filter pruning, and FL was applied as a form of distributed learning.
\par
Another aspect to monitor is the animal health and welfare that can be assessed through observing the animal’s behavior. Animal activity recognition traditionally relies on time-consuming and labor-intensive visual and behavioral observation \cite{muller_new_2003}. Nowadays, it can also be automated through the use of wearable motion data sensors on the animals \citep{mao_fedaar_2022}. The raw data can be interpreted with models trained to classify the behaviors into categories like eating, sleeping, walking etcetera. \cite{mao_fedaar_2022} highlighted that, although FL has the potential of enlarging the size of available data, animal behavior patterns can vary within individuals and between farms leading to client drift, where each client starts to deviate from one another during local training. Furthermore, gradient conflicts might occur during aggregation. Consequently, this work proposed solutions to both problems to encourage convergence towards a global model applicable to all individual animals. \cite{gengler_challenges_2019} proposed utilizing sensor data, usually meant for management applications, to assess and improve the genetic potential of dairy. One of the challenges listed was the governance and ownership of the sensor data. The authors hypothesized that FL would create opportunities for new animal evaluation methods, for instance through training on continuously updated data. The training based on the raw sensor data could result in phenotypic sensor algorithms or even federated genomic prediction models that could enhance breeding research \citep{gengler_challenges_2019}.
\par
Next application addressed the challenges of the monitoring of fermentation processes of beer using ultrasonic sensors and temperature. Due to the length of the fermentation process, it is difficult to collect enough data to train models for each vessel separately, especially for small and medium enterprises. The ultrasonic waves can vary each time a sensor is attached, even if the same procedure is followed and the sensor type and vessel materials stay the same \citep{bowler_transfer_2021}. Each fermentation vessel produces different patterns that are not directly transferable. \cite{bowler_domain_2021} tested domain adaptation strategies to combine data obtained from a laboratory and industrial data (target domain). Federated transfer learning was introduced as a privacy-friendly version of domain adaptation and was compared to combining the data and fine-tuning the models on the target domain.
\par
Other examples of FL for process monitoring can be found in water-related research. The work of \cite{cui_boosted_2022} suggested FL as a future extension of their ground-water level modelling research. FL would be used as a framework for a ground-water level IoT system composed of sensors, edge servers and a central cloud server to obtain up-to-date data. \cite{elhachmi_federated_2022} used this very approach in the context of the water distribution network. They created an FL linear regression model on simulated water that predicts water consumption. Finally, \cite{pei_fed-iout_2023} suggested combining FL with “Internet of Underwater Things” for applications such as water quality monitoring, underwater life detection, flood forecasting, underwater detection, and marine energy transfer. 
\par
\textbf{Crop disease} 
\par
With seven papers on this topic, crop disease identification seems to be the most prevalent use case in the food security domain. Here, FL can be useful to solve challenges related to data sharing such as privacy concerns and communication cost. A large portion of these papers focused on different grains, namely rice \citep{aggarwal_contemporary_2022}, wheat \citep{mehta_empowering_2023}, maize \citep{mehta_revolutionizing_2023}, and the legume soybean \citep{yu_energy-aware_2022}. Two papers were about disease identification in orchards \citep{deng_multiple_2022,tong_detection_2022}. Finally, one paper implemented a system for multiple agricultural pest identification independent of the plant species cultivated \citep{khan_federated_2022}. All of the papers focused on imaging data of healthy and diseased plants. All but the work of \cite{aggarwal_contemporary_2022} implemented an FL system and trained a model for classification or detection.
\par
\cite{khan_federated_2022} and \cite{yu_energy-aware_2022} focused on images obtained thorough unmanned aerial vehicle (UAV) devices. However, only \cite{yu_energy-aware_2022} considered data obtained by drones and collected at the edge-servers. Khan et al. (2022) mentioned UAV technology as the goal, however the experiments were performed on a public dataset without simulating the UAV scenario. \cite{mehta_revolutionizing_2023} mention the problem of specificity and the capabilities of identifying diseases on the individual plant-level being low when remote sensing like drones or satellite images are being utilized. Their solution seemed to focus on the individual-level images of diseased plants. The remaining papers on this topic used openly accessible datasets \citep{mehta_revolutionizing_2023} enriched with images from growers and experts \citep{mehta_empowering_2023}.
\par
\textbf{Food image classification}
\par
A related domain that we chose to include is that of food image classification. The two papers in this category \citep{hsieh_bicameralism_2019,polap_hybridization_2022} were written not from the perspective of solving agricultural problems, instead focusing on advancing the field of FL with food image classification as an example use case. \cite{polap_hybridization_2022} used the fruit image classification problem to showcase an augmentation technique and a new aggregation scheme. \cite{hsieh_bicameralism_2019} introduced a voting scheme that aims to replace regular aggregation (such as averaging the model parameters) in the context of food image classification.
\par
\textbf{Intrusion detection}
\par
The use of IoT technology can lead to new cybersecurity challenges. For instance, an attack targeting a water pump caused a failure of a water treatment plant in Illinois in 2011 \citep{nakashima_foreign_2023}. FL can be applied to ensure the security from malicious cyber-attacks in industrial IoT systems used for monitoring or remote control \citep{jahromi_deep_2021}. The privacy aspect of FL is especially advantageous here, because it allows for the data that we aim to protect to stay at the local client node. Jahromi et al. (2021) presented an FL solution to the detection of cyber security threats in industrial control systems. The usual problem with training such models is that traffic data is sensitive and private, making stakeholders unwilling to share it. \cite{jahromi_deep_2021} used a real-world dataset for Secure Water Treatment in a horizontal federated scenario to train a model for previously unseen attack detection while combining data from different sources. 
\par
In the face of IoT sensors being more frequently employed on farms, it is important to take the safety of those systems into consideration to prevent vulnerabilities to hacker attacks \citep{lu_internet_2019}. Both the work of \cite{kumar_pefl_2022} and \cite{friha_felids_2022} focused on this use case of the protection of agricultural IoT using FL intrusion models.
\cite{bala_novel_2022} highlighted the use of FL and blockchain in the light of crop insurance. Crop insurance requires companies to gather personal data from the farmers. Blockchain as a storage solution is stated to offer transparency and real-time traceability, but at the same time requires a solid consensus strategy to be secure and prevent malicious actors from infiltrating the system. The authors suggest a solution using FL to train a model for calculating trust scores that can be used for a reliable Proof-Of-Stake mechanism for blockchain. 
\par
\textbf{Consumer side of the food supply chain}
\par
The only article we encountered from the consumer side of the food supply chain modelled consumers’ preferences in the context of brand reviews \citep{liu_federated_2022}. In this scenario, each simulated consumer finds different aspects of beer important. Thus, the goal of the collaboration is to reach a personalized solution faster than just modeling each preference separately. FL was implemented so that the consumers can keep their preferences private by only sharing their reward based on the choices made during training, without disclosing which aspects they find important directly. When the consumers’ preferences were modelled using the additional information of collective regret obtained through FL, the authors reached a simulated optimal beer choice faster and with less collective accumulated regret.
\par
\textbf{Federated learning in a broader scope of agriculture digitalization}
\par
FL can be also viewed from the broader perspective of the digitalization of the agriculture. A literature review by \cite{abbasi_digitization_2022} captures and analyzes various developments in smart farming. The digital technologies that were included in the description were the IoT technology, wireless sensor networks, cloud and edge computing, autonomous robotics systems, big data and analytics, ML, deep learning, decision support systems, cyber-physical systems, and digital twins. FL was named as a solution to “digital transformation’s cyber-security and data privacy challenges”. ML, which also includes FL, was considered to be mainly in the prototype phase with few known commercial applications. 

\subsubsection{Application overview: food safety}
This section will zoom in on applications of FL to food safety research. The overview is summarized in Table~\ref{table:applicationsFSafety}. 
\begin{table}[h!]
\resizebox{\textwidth}{!}{%
\begin{tabular}{llll}
\textbf{Domain}              & \textbf{Publication}         & \textbf{Data type}      & \textbf{Data content}                            \\
(mention) Agri-food supply chain & \cite{hassija_survey_2020}       & Unspecified     & Unspecified                             \\
Chemical Hazards         & \cite{yu_food_2022}        & Tabular       & Laboratory results                         \\
Food Fraud            & \cite{gavai_applying_2023}           & Tabular       & Fraud type, product category, year, origin country, control country \\
Quality, milk          & \cite{gulati_beta-fl_2023}           & Imaging       & Ultraviolet visible and near infrared spectrograms         \\
                 & \cite{vimalajeewa_service-based_2022} & Tabular       & Sensor data                             \\
Quality, water          & \cite{park_large-scale_2021}         & Tabular time series & Sensor data                             \\
                 & \cite{vellingiri_strategies_2023}         & Tabular       & Sensor data                             \\
                 & \cite{zhu_research_2021}            & Imaging       & Camera monitoring                          \\
weed detection          & \cite{sharma_weedgan_2022}           & Imaging       & Crop images                             \\
various proposed         & \cite{muller-maatsch_spectral_2021}        & Imaging       & Spectral images                           \\
                 & \cite{qian_perspective_2022}            & Various       &                                   \\
                 & \cite{yaseen_next_2022}           & Imaging,tabular   & Satellite data, in-situ sensor data                
\end{tabular}%
}
\caption{Domain overview food safety research applications of federated learning. }
\label{table:applicationsFSafety}
\end{table}
\par
\textbf{Water quality}
\par
The first group of applications includes water quality assessment and monitoring, described in the papers by \cite{park_large-scale_2021}, Vellingiri et al. (2023), and Zhu (2021). Clean water is not only essential for consumption, but it is also needed for agriculture. If contaminated, irrigation water can be a source of food hazards in crops (Steele \& Odumeru, 2004). That is why essential water sources, such as natural water bodies or water treatment companies, ought to be frequently monitored. The traditional type of water quality monitoring requires more hands-on personnel (Ighalo et al., 2021). Furthermore, some water bodies can be inaccessible and pose a safety risk to the people tasked with obtaining samples (Zhu, 2021). These factors make automating the process beneficial. 
\par
\cite{park_large-scale_2021} tackle the problem of monitoring and prediction of green tide; an overgrowth of algae in water. The data was gathered on the nation-wide scale in South-Korea using smart IoT sensors, edge servers, and a central cloud. The IoT sensors enable monitoring of various green tide indicators such as temperature, pH, and dissolved oxygen. The data cannot feasibly be sent to a central server in real-time, due to the large geographic spread of the sensor locations and the range of the sensor signals being limited. Furthermore, it is not efficient to have a single access point for data collection. This is why multiple local edge servers are used to collect the sensor data. Horizontal FL was applied to use the sensor data on the edge servers to train models locally and aggregate the parameters into a global model. 
\par
Another water quality application focused on predicting pollution of the Cauvery River in India \citep{vellingiri_strategies_2023}. The choice for applying FL was motivated from the data privacy perspective and focused more on enabling collaboration between different parties, in this case water monitoring sites. This data was similar to the data from \cite{park_large-scale_2021}, with the difference of the data being collected at monitoring site-level, making it a cross-silo problem.
\par
The final water quality example by \cite{zhu_research_2021}, employed sensors, in this case cameras, edge servers, and centralized horizontal FL setting. The author uses a convolutional neural network, which is a computer vision model, to assess the quality of the water based on chromatic information in image data. 
\par
\textbf{Milk quality}
\par
FL was also applied to milk quality assessment in liquid \citep{vimalajeewa_service-based_2022} and powder \citep{gulati_beta-fl_2023} form. The milk quality is dependent on its contents and can be analyzed using light spectra from various ranges of light. \cite{vimalajeewa_service-based_2022} used Mid-Infrared Spectroscopy (MIRS) samples for liquid milk quality assessment. This work compares the state-of-the-art techniques to their own composite model in both federated and centralized setting. Interestingly, the first author’s follow up in their PhD thesis was the use of blockchain technology which is a method of securely storing data in a distributed way \citep{vimalajeewa_distributed_2019}. Blockchain technology was proposed to compensate for the drawbacks of centralized FL, namely lack of end-to-end communication, the problem of a single point of failure, the problem with transparency and trustability of the clients and finally the possibility of data leaking through the model updates \citep{vimalajeewa_distributed_2019}. 
\par
The other application assessed milk in the powder form \citep{gulati_beta-fl_2023}. In this work, ultraviolet illumination visible spectral images were collected for both pure and ad-hoc adulterated skimmed milked powder. The author used this case study as a proof-of-concept for combining blockchain technology and FL in supply chains to achieve an FL workflow that ensures protection against malicious attacks on performance from both the server and the clients. Such an FL architecture transfers some of the functionality of a centralized server to a distributed ledger that can choose which local models to aggregate and when to aggregate them based on predefined criteria. 
\par
\textbf{Chemical hazards} 
\par
Pesticides can have adverse health effects on consumers if ingested in amounts exceeding the safety limits \citep{ali_environmental_2021}. Therefore, it is essential to monitor the pesticide residue amount in food to ensure its safety. However, the only publicly available data is often only the pass rate of certain foods \citep{yu_food_2022}. This restricted data availability makes it challenging to conduct research on bioaccumulation of pesticides because products are only described in a qualitative way: under or above the maximum allowed pesticide limit. The actual quantitative values related to the amount of pesticides in food are often scattered across different departments and are not sharable due to data ownership. The only study that makes use of FL in this field is that of \cite{yu_food_2022} in which the aim is to assess the pesticide residue risk in fruits and vegetables in China. To enable the usage of those data silos, the authors created an entropy risk model that can be computed in parts at each local data set and combined at the global server. The resulting model could give a more nuanced view on the pesticide risks of foods compared to the pass rate based on the qualitative values. This FL approach enabled access to raw data that led to more informative metrics, while keeping the data private.
\par
\textbf{Food fraud detection}
\par
FL was also applied to the food safety research on predicting food fraud based on historical data. In their paper, \cite{gavai_applying_2023} simulated a scenario where known food fraud cases for different geographic locations are confidential and not shared beyond their borders. Such experimental setting allowed to combine the food fraud data from the European Union Rapid Alert for Food and Feed (RASFF) database and the United States Economic Motivation Adulteration (EMA). The data was divided among three clients. One client got the data from the EMA database only, and the other two got the data from RASFF split by “before and after 2014”, leading to a realistic data split scenario where data is distributed in a heterogenous way. The authors trained a Bayesian Network on local data only, centralized data, and the local data in FL setting to compare the different scenarios. The main objective of this research was to highlight the potential of FL in the domain of food safety research due to its privacy-preserving character. 
\par
\textbf{Various prospective solutions} 
\par
Finally, across the papers examined in this review, we found various proposals that were only suggested but not implemented by the authors. The perspective review by \cite{muller-maatsch_spectral_2021} suggests the usage of FL for improving the usability of portable miniature spectral devices for measuring food safety, authenticity, and quality. The challenge they addressed is the large bottleneck caused by the need to calibrate and re-calibrate such instruments; a task for which the necessary databases are often device specific, food specific, or contamination specific. Due to such heterogeneity across these datasets, eventual similarities between databases are not being leveraged to their full potential and, in consequence, every change is met with a lot of repetitive work. AN FL network was proposed to show how a data sample can be linked to different data sources such as reference methods database, spectral imaging database, and more. 
\par
Another perspective paper \citep{yaseen_next_2022} proposed using FL together with an edge cloud server to detect and predict heavy metal contamination of soil and water. FL was presented as a way of facilitating the use of various data types, satellite imagery in combination of in-situ sampling data, gathered at the edges of the edge cloud server while keeping the sources private.
\par
Finally, a review paper by \cite{qian_perspective_2022} proposed a wide range of potential application domains for FL and other private data sharing methods. The authors named multiple kinds of food safety data with different degrees of sharing concerns: microbial data, environmental data including risk factors, processing and management data, consumer data, and supply chain data. They also suggested model types that may benefit from private data sharing strategies, including but not limited to FL. Quantitative microbial risk assessment models for, among others, the prediction of foodborne illness or recall likelihoods, often combine multiple data sources. Adding more data sources can decrease the uncertainty of the predictions and improve the validation process \citep{qian_perspective_2022}. Graphic information system (GIS) models often use a combination of microbial data, and both private and open-access environmental data. Microbial data is often costly to gather, and the models would greatly benefit from sharing those datasets. For instance, it would both improve the spatial and temporal distribution of the data \citep{qian_perspective_2022}. Agent-based models are models simulating complex systems through simulation of autonomous components. While they are often facility specific, data sharing could help with publication of the results, improving the parameter estimates, and creation of more generic agent-based models \citep{qian_perspective_2022}. Lastly, public health models for foodborne disease outbreaks often combine genomic data and consumer data. Consumer data can improve upon the addition of more data varieties, like data from e-commerce, increased spatial and temporal dimensions, and the specification of bias in the collected data by sharing the demographic information \citep{qian_perspective_2022}. 
\par
\textbf{Bordering on food safety}
\par
We decided to include two works not completely meeting our relevance criteria to involve both computer-vision \citep{sharma_weedgan_2022} and supply chain security \citep{hassija_survey_2020}) cases into the review because deemed relevant as example of future application of FL in food safety.
\par
First, \cite{sharma_weedgan_2022} used FL to obtain a better weed image generator to train better weed identification algorithms with the goal of herbicide usage reduction. Herbicide used on crops can be harmful to both people working in the fields and ingesting the products. This work considered two separately trained discriminators to be client nodes, that receive equally sized random batches of data during training. Since there are no constant local datasets that are strictly separated, calling this approach FL is questionable. 
\par
\cite{hassija_survey_2020} suggested the use of FL in supply chains as a new privacy-preserving ML technique. The authors discussed various security-critical supply chain areas, including the agri-food supply chain. They emphasized the strong requirement for safety measures in this specific supply chain that is needed for guarding the product safety. Early warning systems for food suppliers were suggested as a solution to limit risks in the food supply chain.

\subsubsection{Federated learning performance in the food domain}
Some papers were purely perspective papers and proposed future directions of FL \citep{qian_perspective_2022, hassija_survey_2020, muller-maatsch_spectral_2021, yaseen_next_2022}. For the papers that implemented and tested an FL system, the question remains how well the research setup matches the problems of FL applied to data collected by different organizations or devices. One of the challenges of FL is non-independent and identically distributed (non-IID) data \citep{li_review_2020}. The data across clients is not guaranteed to have the same distribution, for instance due to differences amongst farms. The lack of guarantees for data homogeneity can introduce problems with model convergence and performance \citep{zhao_federated_2018}. However, this potential problem is not addressed when the data is artificially partitioned into equal, balanced sets. Many of the authors are not explicit in their specification of the data division method \citep{jahromi_deep_2021,vellingiri_strategies_2023,vimalajeewa_service-based_2022,zhu_research_2021}, or partition the data in an optimistic, uniformly distributed way \citep{gulati_beta-fl_2023,huang_high-precision_2022,park_large-scale_2021,sharma_weedgan_2022}. 
\par
\cite{jahromi_deep_2021} concluded that an FL anomaly detection method outperforms non-FL methods with only partial access to the data, solves privacy and security issues bound to sharing data directly and circumvents the time-consuming training problem on a single centralized server by performing the training in a decentralized way. However, their results did not consider the non-IID data problem, and the models were not compared to a model trained on the combined data. The latter is useful to assess how well FL performs against the optimal case where all data is shared directly. \cite{vellingiri_strategies_2023} achieved a more accurate performance and reduced training time using the FL approach compared to the individually trained local models.Unfortunately, it was not possible to evaluate the advantages of FL over non-FL because the author did not evaluate the performance of a non-FL approach as well. 
\par
The work of \cite{vimalajeewa_service-based_2022} found that they could achieve a performance similar to that of the state-of-the-art convolutional neural network (CNN) model for milk classification under the FL setting. \cite{zhu_research_2021} reported that the FL models reached the same performance as a conventional CNN faster, while also reducing the human and material resource requirements. However, the author did not release any results to back-up these claims, making the results less reliable due to the lack of transparency. \cite{gulati_beta-fl_2023} compared their Blockchain-Event Triggered Asynchronous FL scheme with another synchronous FL system, and synchronous FL on the Blockchain. Their FL communication scheme achieved similar performance but with more claims to protection against malicious agents. Similarly, \cite{park_large-scale_2021} compared different scheduling scenarios within the FL framework. The authors concluded that their scheduling method achieves effective performance due to balanced data gathering. However, in their simulation, they did not consider real-world factors like positions of the network components, distribution ratios and communication channel qualities. The WeedGAN model developed by \cite{sharma_weedgan_2022} used FL in the form of two distributed discriminator models to train a generative adversarial network for creating synthetic cotton weed images. Seven state-of-the-art fine-tuned image classification models trained on this synthetic dataset outperformed models trained on the raw data and the augmented data. 
\par
Some papers reflected the non-IID data problem by dividing the data amongst the artificial clients in a realistic manner. The data was divided based on the geographical location \citep{durrant_role_2022,gavai_applying_2023,yu_food_2022} or time \cite{gavai_applying_2023}. The two experiments performed by \cite{gavai_applying_2023} showed that FL setting did not substantially decrease the performance when compared to a centralized setting. The difference between locally trained models and FL was not straightforward, because in some cases the local performance decreased. The authors concluded that it was most likely due to data heterogeneity. One client did not contain some of the possible food fraud cases, and enriching the model with those cases made its local performance worse. However, the authors also claim that the model created in the FL setting contains more information and is more realistic than the local models, because it has access to more data. 
\par
\cite{yu_food_2022} did not quantify the performance of their federated setting entropy model of pesticides residues, making it impossible to assess the model in terms of metrics. However, qualitatively they claim that their FL entropy model is more informative than just the summary metrics. 
\par
Finally, \cite{huang_high-precision_2022} performed a detailed analysis testing many scenarios, including a varying number of clients, different data division, lightweight model versus original model comparison, and FL or no FL comparison. The accuracy of models using FL was consistently better and the difference in performance between the lightweight version and the original version of the models seemed to be smaller in the federated setting. The different computer vision models tested seemed to have a varying degree of robustness against various data partitioning settings.

\subsubsection{Reasons for federated learning}
The main reasons listed by the included literature to use FL can be categorized into: using FL to train models on local data that otherwise would not be shared, and using FL as a form of efficient resource allocation. In this context, we define resource limitations to be any limitations concerning computing power, storage place, or communication (i.e., latency or overload). Efficient resource allocation aims to counter these problems.
\par
The first argument encompasses the interconnected concepts of data sharing \citep{gavai_applying_2023,gulati_beta-fl_2023,muller-maatsch_spectral_2021,qian_perspective_2022,yaseen_next_2022,yu_food_2022,yu_energy-aware_2022}, retaining privacy \citep{gavai_applying_2023,gulati_beta-fl_2023, muller-maatsch_spectral_2021,qian_perspective_2022,vellingiri_strategies_2023,vimalajeewa_service-based_2022,yaseen_next_2022}, data access rights and confidentiality \citep{gavai_applying_2023,muller-maatsch_spectral_2021,yu_food_2022}, security \citep{gavai_applying_2023}, and data ownership \cite{gavai_applying_2023,vimalajeewa_service-based_2022}. Some combination of these terms was provided as a reason in many of the encountered papers \citep{durrant_role_2022,elhachmi_federated_2022,friha_felids_2022,gengler_challenges_2019,idoje_federated_2023,khan_federated_2022,kumar_pefl_2022,manoj_federated_2022,mao_fedaar_2022,mehta_empowering_2023,mehta_revolutionizing_2023,pei_fed-iout_2023,tong_detection_2022, zhang_maize_2023}. 
\par
The argument of resource allocation is often encountered in edge computing scenarios \citep{deng_multiple_2022,elhachmi_federated_2022,friha_felids_2022,gengler_challenges_2019,huang_high-precision_2022, idoje_federated_2023,khan_federated_2022,liu_federated_2022,pei_fed-iout_2023,tong_detection_2022,yu_energy-aware_2022}. For instance, FL is useful in cases where the data is geographically scattered and resides on devices with a weak signal range. This might be the case with IoT technology where FL is posed as a solution to train the models in a feasible and robust manner on the edge servers that collect the data from nearby devices. In another instance, \cite{park_large-scale_2021} found FL to be a robust solution in a setting with online data gathering, where communicating the data to a central server was not feasible. Other problems concerning resources included centralized IoT communication, resource management, limitation to transmission distances, burdens of massive accesses, limitations of ICT infrastructure, and limitations on personnel \citep{park_large-scale_2021,sharma_weedgan_2022,vimalajeewa_distributed_2019,zhu_research_2021}.
\par
Two papers had the advancement of the FL field as their main goal, while using food-related data to benchmark their research \citep{hsieh_bicameralism_2019,polap_hybridization_2022}.

\subsubsection{Food applications classified to the federated learning framework}
Table~\ref{table:flfood} and Table~\ref{table:flfs} summarize how the current and proposed applications within food and food safety span the FL landscape.
\begin{table}[]
\resizebox{\textwidth}{!}{%
\begin{tabular}{
>{\columncolor[HTML]{FFFFFF}}l 
>{\columncolor[HTML]{FFFFFF}}l 
>{\columncolor[HTML]{FFFFFF}}l 
>{\columncolor[HTML]{FFFFFF}}l 
>{\columncolor[HTML]{FFFFFF}}l }
\textbf{Reference}            & \textbf{Architecture}     & \multicolumn{2}{l}{\cellcolor[HTML]{FFFFFF}\textbf{Data Partition}}     & \textbf{Cross-silo / cross-device}   \\
\cite{abbasi_digitization_2022}      & \multicolumn{2}{l}{\cellcolor[HTML]{FFFFFF}None}    & None                 & None             \\
\cite{aggarwal_contemporary_2022}     & \multicolumn{2}{l}{\cellcolor[HTML]{FFFFFF}Centralized} & Unspecified              & Cross-device         \\
\cite{bala_novel_2022}      & \multicolumn{2}{l}{\cellcolor[HTML]{FFFFFF}Centralized} & Horizontal              & Cross-silo          \\
\cite{bera_e-cropreco_2023}       & \multicolumn{2}{l}{\cellcolor[HTML]{FFFFFF}Centralized} & Horizontal              & Cross-silo          \\
\cite{bowler_domain_2021}      & \multicolumn{2}{l}{\cellcolor[HTML]{FFFFFF}Centralized} & Horizontal              & Cross-silo          \\
\cite{cui_boosted_2022}       & \multicolumn{2}{l}{\cellcolor[HTML]{FFFFFF}Centralized} & Horizontal              & Cross-device         \\
\cite{deng_multiple_2022}       & \multicolumn{2}{l}{\cellcolor[HTML]{FFFFFF}Centralized} & Horizontal              & Cross-silo          \\
\cite{durrant_role_2022}      & \multicolumn{2}{l}{\cellcolor[HTML]{FFFFFF}Centralized} & Horizontal              & Cross-silo          \\
\cite{elhachmi_federated_2022}   & \multicolumn{2}{l}{\cellcolor[HTML]{FFFFFF}Centralized} & Unspecified              & Cross-device         \\
\cite{friha_felids_2022}       & \multicolumn{2}{l}{\cellcolor[HTML]{FFFFFF}Centralized} & Horizontal              & Cross-silo  or cross-device \\
\cite{gengler_challenges_2019}         & \multicolumn{2}{l}{\cellcolor[HTML]{FFFFFF}Unspecified} & Unspecified              & Unspecified         \\
\cite{hsieh_bicameralism_2019}      & \multicolumn{2}{l}{\cellcolor[HTML]{FFFFFF}Centralized} & Transfer  learning          & Cross-device         \\
\cite{huang_high-precision_2022}     & \multicolumn{2}{l}{\cellcolor[HTML]{FFFFFF}Centralized} & Horizontal              & Cross-device         \\
\cite{idoje_federated_2023}      & \multicolumn{2}{l}{\cellcolor[HTML]{FFFFFF}Centralized} & Horizontal              & Cross-silo          \\
\cite{khan_federated_2022}       & \multicolumn{2}{l}{\cellcolor[HTML]{FFFFFF}Centralized} & Horizontal              & Cross-silo or cross device  \\
\cite{kumar_pefl_2022}     & \multicolumn{2}{l}{\cellcolor[HTML]{FFFFFF}Centralized} & Transfer  learning          & Cross-silo  or cross device \\
\cite{liu_federated_2022}      & \multicolumn{2}{l}{\cellcolor[HTML]{FFFFFF}Centralized} & Transfer learning           & Cross-silo          \\
\cite{manoj_federated_2022}       & \multicolumn{2}{l}{\cellcolor[HTML]{FFFFFF}Centralized} & Horizontal              & Cross-silo          \\
\cite{mao_fedaar_2022}       & \multicolumn{2}{l}{\cellcolor[HTML]{FFFFFF}Centralized} & Horizontal              & Cross-silo          \\
\cite{mehta_empowering_2023}& \multicolumn{2}{l}{\cellcolor[HTML]{FFFFFF}Centralized} & Horizontal              & Cross-silo          \\
\cite{mehta_revolutionizing_2023} & \multicolumn{2}{l}{\cellcolor[HTML]{FFFFFF}Centralized} & Horizontal              & Cross-silo          \\
\cite{pei_fed-iout_2023}       & \multicolumn{2}{l}{\cellcolor[HTML]{FFFFFF}Centralized} & Unspecified              & Cross-device         \\
\cite{polap_hybridization_2022}     & \multicolumn{2}{l}{\cellcolor[HTML]{FFFFFF}Centralized} & Horizontal              & Cross-silo          \\
\cite{tong_detection_2022}       & \multicolumn{2}{l}{\cellcolor[HTML]{FFFFFF}Centralized} & Horizontal              & Cross-silo          \\
\cite{yu_energy-aware_2022}      & \multicolumn{2}{l}{\cellcolor[HTML]{FFFFFF}Centralized} & Horizontal              & Cross-device         \\
\cite{zhang_maize_2023}     & \multicolumn{2}{l}{\cellcolor[HTML]{FFFFFF}Centralized} & Vertical               & Cross-silo         
\end{tabular}%
}
\caption{Federated learning framework description of the literature in food}
\label{table:flfood}
\end{table}
\begin{table}[]
\resizebox{\textwidth}{!}{%
\begin{tabular}{
>{\columncolor[HTML]{FFFFFF}}l 
>{\columncolor[HTML]{FFFFFF}}l 
>{\columncolor[HTML]{FFFFFF}}l 
>{\columncolor[HTML]{FFFFFF}}l }
\textbf{Reference}           & \textbf{Architecture}        & \textbf{Data Partition}   & \textbf{Cross-silo / cross-device} \\
\cite{gavai_applying_2023}     & Centralized         & Horizontal     & Cross-silo        \\
\cite{gulati_beta-fl_2023}     & Centralized/(decentralized) & Horizontal     & Cross-silo        \\
\cite{muller-maatsch_spectral_2021} & Unspecified         & Horizontal/Vertical & Unspecified        \\
\cite{park_large-scale_2021}      & Centralized         & Horizontal     & Cross-device       \\
\cite{sharma_weedgan_2022}     & Centralized         & Horizontal     & Cross-silo        \\
\cite{vellingiri_strategies_2023}   & Centralized         & Horizontal     & Cross-silo        \\
\cite{vimalajeewa_distributed_2019}      & Centralized         & Horizontal     & Cross-silo/cross-device  \\
\cite{yu_food_2022}     & Centralized         & Horizontal     & Cross-silo        \\
\cite{zhu_research_2021}         & Centralized         & Horizontal     & Cross-device       \\
\cite{yaseen_next_2022}        & Centralized         & Horizontal/Vertical & Cross-silo/cross-device  \\
\cite{jahromi_deep_2021}   & Centralized         & Horizontal     & Cross-silo        \\
\cite{huang_high-precision_2022}    & Centralized         & Horizontal     & Cross-device       
\end{tabular}%
}
\caption{Federated learning framework description of the literature in food safety}
\label{table:flfs}
\end{table}
\par
All papers found on this topic either used a centralized architecture or did not specify the architecture. The exception was arguably \cite{gulati_beta-fl_2023} where the authors combined centralized FL architecture with a block-chain scheme that made the setting more decentralized. 
\par
Most applications and propositions focused on horizontal partitioning. \cite{muller-maatsch_spectral_2021} and \cite{yaseen_next_2022} were the exception in food safety. \cite{yaseen_next_2022} proposes combining multiple data sources at edge servers. However, it is unclear whether each edge server would have access to the same type of sensor data and satellite data. Similarly, \cite{muller-maatsch_spectral_2021} were unclear in their proposed federated learning application. \cite{zhang_maize_2023} had the only application explicitly focusing on the vertical FL where the client nodes did not have access to the same set of features. Some applications included transfer learning elements, such as the work of \cite{hsieh_bicameralism_2019}. In their work, \cite{hsieh_bicameralism_2019} pretrain the model on a source task and fine-tune it on the target task.
\par
The cross-silo and cross-device split was difficult to pinpoint at times. For instance, one paper was aimed at what seemed to be a cross-device scenario, while the experimental setting was effectively cross-silo where the sensor data was divided across five clients \citep{vimalajeewa_distributed_2019}. \cite{gulati_beta-fl_2023} did not specify the number of clients, but it is implied that each client is supposed to represent an organization. \cite{vellingiri_strategies_2023} mentioned sixteen monitoring sites that posed as client nodes, which lead to the assumption that the experimental setting was composed of sixteen data silos. \cite{park_large-scale_2021} simulated the scheduling of IoT sensors at the edge servers within the federated learning task, bringing it closer to the cross-device scenario. \cite{huang_high-precision_2022} was also classified as cross-device. Although their number of participating clients was low (5), they were heterogeneous and contained mobile phones, reflecting the hardware heterogeneity aspect of this setting. 

\section{Discussion}
We have reviewed and discussed 41 papers using federated learning (FL) in the food domain. From this, several trends and potential gaps can be identified. 
\par
Surprisingly, almost no paper implemented a vertical partitioning of the data. Vertical partitioning is useful when one wants to combine different features on the same entity scattered across multiple sources. As compared to horizontal FL, the benefit of using vertical FL includes: 1) each party holds different subsets of features, which allows stakeholders to contribute specific domain knowledge and expertise, leading to a more comprehensive and nuanced understanding of the problem; and 2) by focusing on relevant features from different stages of the food supply chain, vertical FL can reduce redundancy in data transmission and processing. This efficiency in resource utilization can lead to faster convergence during model training and lower communication overhead. Such an application could be useful for improving the safety of the food supply chains. In this context, the safety of a product is influenced by its production, processing, transportation, storage, and consumption, and the associated data are owned by the company handling each of these steps. Consequently, FL could facilitate collaboration between the companies involved in the supply chain, connect various aspects of food processing and handling into one model for the early warning system of the food supply chain.
\par
New methods that improve traceability of the products, such as blockchain \citep{hassija_survey_2020}, could be combined with FL. Some publications we encountered already combined the two technologies \citep{bala_novel_2022,gulati_beta-fl_2023}, but not yet in a vertical data partitioning scenario. The reason behind this uncommon implementation of vertical FL is likely due to some exclusive challenges typical of this type of FL. For example, sharing common identifiers or features across vertical partitions poses privacy risks, as these attributes may inadvertently reveal sensitive information about individuals, products, or processes. Safeguarding against attribute linkage attacks requires careful consideration of privacy-preserving techniques and encryption methods \citep{liu_vertical_2023}. 
\par
We also found a lack of federated transfer learning research. Besides preserving data privacy and confidentiality, the benefits of using federated transfer learning research could be improved model performance. Federated transfer learning provides a head start by initializing the model with knowledge learned from a large, diverse, or available dataset. Fine-tuning with FL allows the model to adapt to local data distributions and domain-specific features, which may lead to improved performance and generalization across different food safety scenarios. An example research direction could be the transfer of experimental knowledge to industrial settings. As we saw in the paper by \cite{bowler_domain_2021}, FL can be used as a domain adaptation method allowing a model to be pretrained on a small scale experimental setting, and adapted to a larger industrial setting. Their study used previously collected ultrasonic sensor data from laboratory scale fermentations to improve the model prediction accuracy on an industrial scale fermentation process. Their result showed the federated transfer learning methodology achieved higher accuracy for 14 out of 16 ML tasks compared to the baseline model. Similar domain adaptation methodology may be applied in other domains that require monitoring. 
\par
The server architectures were almost exclusively centralized, with the exception of \cite{gulati_beta-fl_2023} who combined FL with a distributed blockchain ledger. As compared to centralized FL, decentralized FL improves scalability by distributing computational tasks and communication overhead across multiple client nodes, enabling FL to scale to larger networks of participants. Furthermore, decentralized FL architectures may be considered in cases where having a single point of failure is a drawback or the communication with the centralized server is a bottleneck. In food safety domain, decentralized FL could be more suitable for large-scale food safety networks. However, decentralized FL also comes with its own challenges. For instance, decentralized FL may be more prone to cybersecurity problems \citep{martinez_beltran_decentralized_2023}.
\par
The future employment of IoT technology in food production and monitoring may lead to new data sources suitable for the cross-device FL scenario. We already encountered examples of animal behavior sensors \citep{gengler_challenges_2019,mao_fedaar_2022}, ultrasonic sensors \citep{bowler_domain_2021}, and remote sensing technology like UAV \citep{khan_federated_2022,yu_energy-aware_2022} paired with FL. IoT sensors, such as temperature and humidity sensors, can potentially be used in food safety research \citep{bouzembrak_internet_2019}. For now, most of the IoT-based agricultural systems seem to be still in the prototypical or conceptual stage \citep{abbasi_digitization_2022}. With the increased availability of data, food safety research could leverage large-scale datasets to identify patterns, trends, and correlations that can inform risk assessment, hazard analysis, forecasting food safety risks, and identifying emerging threats and mitigation strategies.
\par
Finally, during the review emerged that FL does not guarantee full protection of the data or the fairness of the models. Malicious agents may use the parameter updates to extract sensitive information about the data or disrupt the model training \citep{lyu_privacy_2022}. FL can be made safer with measures such as differential privacy or encryption. Such measures were considered in few of the found papers, for example the work of \cite{durrant_role_2022}. Official guidelines for best practices in FL are needed for responsible and safe use of local data. The European Union has guidelines for artificial intelligence (AI) models described in sources such as the AI Act and best practices on securing ML algorithms provided by European Union Agency for Cybersecurity (ENISA) \citep{malatras_securing_2021}. ENISA also provides guidelines on data-sharing and (pseudo)anonymization. However, to our knowledge, no official policies exist within the European Union that would directly address security and privacy standards of FL applications.

\section{Conclusion}
This review collected and summarized all publications of FL applications in the food safety domain till August 2023. Despite the benefits FL has to offer, the number of papers currently available was limited. As of today, the food safety applications of FL include water quality, milk quality, chemical hazards, food fraud, and some prospective solutions. Other FL applications were classified into yield prediction, process monitoring and optimization, crop disease prediction and recognition, classification of foods, IoT intrusion detection, and consumer preference. There were two main reasons for the employment of FL found in the literature: data sharing-related and resource-related reasons. The main FL trend in food was the use of horizontal, centralized FL. The pattern of the type of client nodes (cross-silo vs cross-device) was less clear. Comparing the current applications to the FL framework showcased some gaps, such as the lack of vertical data partitioning, federated transfer learning applications, and decentralized architectures. Incorporating vertical data partitioning would enable the integration of complementary datasets from different stages of the food supply chain, facilitating comprehensive risk assessment and mitigation strategies. Federated transfer learning applications could leverage pretrained models on related domains, accelerating the development of accurate food safety prediction models with limited labeled data. Introducing decentralized architectures would enable larger numbers of client nodes thanks to the decrease of communication overhead. Furthermore, the broader look on the food domain showed some trends for the use of FL alongside IoT technology and sensors. FL can then leverage this wealth of diverse and high-dimensional data to develop predictive models for identifying potential food safety hazards, optimizing storage and transportation conditions, and implementing proactive risk management strategies. We anticipate this paper to be useful for researchers and companies that aim to use ML on food-related data that previously could not be shared due to privacy concerns or want to have a better overview of FL in food research.

\newpage
\section*{Acknowledgments}
We would like to thank Mary Godec for her insights and advice.

\section*{Funding}
Funding for this research has been provided by the European Union’s Horizon Europe research and innovation programme EFRA [grant number 101093026]. 
\par
This project has received funding from the European Union’s Horizon Europe research and innovation programme as a part of the HOLiFOOD project [grant number 101059813].
Funded by the European Union. 
\par
Funded by the European Union. 
\par
This project has also been co-funded by the Netherlands Ministry of Agriculture, Nature, and Food Quality (LNV) through TKI.
\par
Views and opinions expressed are however those of the author(s) only and do not necessarily reflect those of the European Union or Research Executive Agency. Neither the European Union nor the granting authority can be held responsible for them.

\bibliographystyle{apacite}
\bibliography{references} 

\section*{Appendix}
Search codes for literature search:
The search was performed on the title, the abstract and the keywords. We obtained the articles using following search terms on 31st of July 2023.
\par
\textbf{Web of science}
\par
TS=("federated learning" OR "federated approach") 
AND (TS=(food* AND (safety OR monitoring OR supply OR chain OR processing OR market* OR distribution OR storage OR handling OR farm* OR fraud OR hygiene OR surveillance OR agribusiness)) OR TS=(food AND (adulteration OR contaminants OR tempering OR fraud)) OR TS=(water OR food* OR vegetable* OR fruit* OR meat OR fish OR grain OR cereal OR legume* OR milk OR dairy OR egg* OR maize OR rice OR wheat OR flour OR potato* OR onion* OR tomato* OR lettuce OR carrot* OR pepper* OR cucumber* OR celery OR broccoli OR mushroom OR spinach OR cabbage* OR bean* OR cauliflower OR garlic OR asparagus OR banana* OR apple* OR grape* OR strawberries OR melon* OR avocado* OR blueberries OR mandarins OR oranges OR peach* OR pineapple* OR lemon* OR poultry OR chicken* OR pig* OR duck* OR bovine OR beef OR broiler OR salmon OR tuna OR tilapia OR bacteria OR "Bacillus cereus" OR "Campylobacter jejuni" OR "Clostridium botulinum" OR "Clostridium perfringens" OR "e. coli" OR "Escherichia coli" OR "Listeria monocytogenes" OR Shigella OR "Staphylococcus aureus" OR salmonella OR "Vibrio cholerae" OR "Vibrio parahaemolyticus" OR "Vibrio vulnificus" OR "Yersinia enterocolitica" OR "acromobacter sakazakii" OR "Bacteriophage" OR "Enteric Virus" OR "Hepatitis A virus" OR Norovirus OR "Norwalk virus" OR "Rota virus" OR prion OR "Mad Cow Disease" OR "food poisoning" OR fungus OR mycotoxin* OR aflatoxin OR Deoxynivalenol OR "Ochratoxin A" OR Fumonisin OR Patulin OR arsenic OR cadmium OR Mercury OR "heavy metals" OR pesticide* OR pest* OR instecticide* OR fungicide* OR herbicide* OR Azoxystrobin OR Captan OR Clethodim OR Thiocarbamate OR "allergenic" OR "Anaphylactic shock" OR "allergen*") OR TS=(crop* OR agriculture))
NOT TS=(“fine-grained” OR cancer)
\par
\textbf{Google scholar}
\par
("federated learning" OR "federated approach") AND (food AND (safety OR monitoring OR supply OR chain OR processing OR market* OR distribution OR storage OR handling OR farm* OR fraud OR hygiene OR surveillance OR agribusiness ) OR (food AND (adulteration OR contaminants OR tempering OR fraud) OR bacteria OR “Bacillus cereus” OR “Campylobacter jejuni” OR “Clostridium botulinum” OR “Clostridium perfringens” OR "e. coli" OR "Escherichia coli" OR ”Listeria monocytogenes” OR Shigella OR “Staphylococcus aureus” OR salmonella OR “Vibrio cholerae” OR “Vibrio parahaemolyticus” OR “Vibrio vulnificus” OR “Yersinia enterocolitica” OR “Cronobacter sakazakii” OR “Bacteriophage” OR “Enteric Virus” OR “Hepatitis A” virus OR Norovirus OR “Norwalk virus” OR “Rota virus” OR fungus OR mycotoxin* OR pesticide* OR pest* OR vegetable* OR fruit* OR meat OR fish OR grain OR cereal OR legume* OR milk OR dairy OR egg* OR crop* OR agriculture)
\par
\textbf{CAB abstracts}
\par
CAB abstracts is a database already limited to articles on agriculture and life sciences. Since using just federated learning and federated approach as search terms resulted in only 15 hits, we decided not to narrow it down any more than that.
\par
("federated learning" OR "federated approach")
\par
\textbf{Scopus (126 hits total):}
\par
TITLE-ABS-KEY ( "federated learning" OR "federated approach" ) AND ( TITLE-ABS-KEY ( food* AND ( safety OR monitoring OR supply OR chain OR processing OR market* OR distribution OR storage OR handling OR farm* OR fraud OR hygiene OR surveillance OR agribusiness ) ) OR TITLE-ABS-KEY ( food AND ( adulteration OR contaminants OR tempering OR fraud ) ) OR TITLE-ABS-KEY ( water OR food* OR vegetable* OR fruit* OR meat OR fish OR grain OR cereal OR legume* OR milk OR dairy OR egg* OR maize OR rice OR wheat OR flour OR potato* OR onion* OR tomato* OR lettuce OR carrot* OR pepper* OR cucumber* OR celery OR broccoli OR mushroom OR spinach OR cabbage* OR bean* OR cauliflower OR garlic OR asparagus OR banana* OR apple* OR grape* OR strawberries OR melon* OR avocado* OR blueberries OR mandarins OR oranges OR peach* OR pineapple* OR lemon* OR poultry OR chicken* OR pig* OR duck* OR bovine OR beef OR broiler OR salmon OR tuna OR tilapia OR bacteria OR "Bacillus cereus" OR "Campylobacter jejuni" OR "Clostridium botulinum" OR "Clostridium perfringens" OR "e. coli" OR "Escherichia coli" OR "Listeria monocytogenes" OR shigella OR "Staphylococcus aureus" OR salmonella OR "Vibrio cholerae" OR "Vibrio parahaemolyticus" OR "Vibrio vulnificus" OR "Yersinia enterocolitica" OR "acromobacter sakazakii" OR "Bacteriophage" OR "Enteric Virus" OR "Hepatitis A virus" OR norovirus OR "Norwalk virus" OR "Rota virus" OR prion OR "Mad Cow Disease" OR "food poisoning" OR fungus OR mycotoxin* OR aflatoxin OR deoxynivalenol OR "Ochratoxin A" OR fumonisin OR patulin OR arsenic OR cadmium OR mercury OR "heavy metals" OR pesticide* OR pest* OR instecticide* OR fungicide* OR herbicide* OR azoxystrobin OR captan OR clethodim OR thiocarbamate OR "allergenic" OR "Anaphylactic shock" OR "allergen*" ) OR TITLE-ABS-KEY ( crop* OR agriculture ) ) AND NOT TITLE-ABS-KEY ( "fine-grained" OR cancer )
\par
\textbf{IEEE divided into: (83 hits total)}
\par
1. ( "federated learning" OR "federated approach") AND ( food* AND ( safety OR monitoring OR supply OR chain OR processing OR market* OR distribution OR storage OR handling OR farm* OR fraud OR hygiene OR surveillance OR agribusiness ) ) 15 hits
 \par
 2. ( "federated learning" OR "federated approach") AND ( food AND ( adulteration OR contaminants OR tempering OR fraud ) ) 
\par
3. ("federated learning" OR "federated approach") AND (water OR food* OR vegetable* OR fruit* OR meat OR fish OR grain OR cereal OR legume* OR milk OR dairy OR egg* OR maize OR rice) NOT (FINE-grained OR cancer) 63 hits
\par
4. ("federated learning" OR "federated approach") AND (OR wheat OR flour OR potato* OR onion* OR tomato* OR lettuce OR carrot* OR pepper* OR cucumber* OR celery OR broccoli OR mushroom OR spinach OR cabbage* OR bean* OR cauliflower OR garlic OR asparagus 
\par
5. ("federated learning" OR "federated approach") AND (banana* OR apple* OR grape* OR strawberries OR melon* OR avocado* OR blueberries OR mandarins OR oranges OR peach* OR pineapple* OR lemon*) 
6. ("federated learning" OR "federated approach") AND (poultry OR chicken* OR pig* OR duck* OR bovine OR beef OR broiler OR salmon OR tuna OR tilapia)
\par
7. ("federated learning" OR "federated approach") AND (bacteria OR "Bacillus cereus" OR "Campylobacter jejuni" OR "Clostridium botulinum" OR "Clostridium perfringens" OR "e. coli" OR "Escherichia coli" OR "Listeria monocytogenes" OR Shigella OR "Staphylococcus aureus" OR salmonella OR "Vibrio cholerae" OR "Vibrio parahaemolyticus" OR "Vibrio vulnificus")
\par
8. ("federated learning" OR "federated approach") AND (OR "Yersinia enterocolitica" OR "acromobacter sakazakii" OR "Bacteriophage" OR "Enteric Virus" OR "Hepatitis A virus" OR Norovirus OR "Norwalk virus" OR "Rota virus" OR prion OR "Mad Cow Disease" OR "food poisoning")
\par
9. ("federated learning" OR "federated approach") AND (fungus OR mycotoxin* OR aflatoxin OR Deoxynivalenol OR "Ochratoxin A" OR Fumonisin OR Patulin OR arsenic OR cadmium OR Mercury OR "heavy metals" OR pesticide* OR pest*) 5 hits
\par
10. ("federated learning" OR "federated approach") AND (instecticide* OR fungicide* OR herbicide* OR Azoxystrobin OR Captan OR Clethodim OR Thiocarbamate OR "allergenic" OR "Anaphylactic shock" OR "allergen*")
\end{document}